\pdfoutput=1

\documentclass[11pt]{article}

\usepackage[]{acl}

\usepackage{times}
\usepackage{latexsym}

\usepackage{algorithmic}

\usepackage{latexsym}

\usepackage{colortbl}
\usepackage{booktabs}
\usepackage{multirow}
\usepackage{graphicx}
\usepackage{subcaption}
\usepackage{pgfplots}
\usepackage[ruled,vlined]{algorithm2e}
\usepackage[T1]{fontenc}
\usepackage{diagbox}

\usepackage[T1]{fontenc}

\usepackage[utf8]{inputenc}

\usepackage{microtype}

\title{Sarcasm Detection in a Disaster Context}

 \author{
    \textbf{Tiberiu Sosea}$^1$\quad\textbf{Junyi Jessy Li}$^2$\quad\textbf{Cornelia Caragea}$^1$\\
    $^1$Department of Computer Science, University of Illinois Chicago\\$^2$Department of Linguistics, The University of Texas at Austin\\
    {\color{blue}\texttt{\{tsosea2,cornelia\}@uic.edu}\quad\texttt{\{jessy\}@utexas.edu}}
}

\begin{document}
\maketitle
\begin{abstract}
During natural disasters, people often use social media platforms such as Twitter to ask for help, to provide information about the disaster situation, or to express contempt about the unfolding event or public policies and guidelines. 
This contempt is in some cases expressed as sarcasm or irony. Understanding this form of speech in a disaster-centric context is essential to improving natural language understanding of disaster-related tweets. 
In this paper, we introduce \textsc{HurricaneSARC}, a dataset of $15,000$ tweets annotated for intended sarcasm, and provide a comprehensive investigation of sarcasm detection using pre-trained language models. Our best model is able to obtain as much as $0.70$ F1 on our dataset. 
We also demonstrate that 
the performance on \textsc{HurricaneSARC} can be improved 
by
leveraging intermediate task transfer learning. We release our data and code at \url{https://github.com/tsosea2/HurricaneSarc}.
\end{abstract}

\section{Introduction}

Understanding sarcasm from text is crucial for enabling progress on natural language understanding.
However, despite being widely researched as an NLP task \cite{riloff2013sarcasm,singh2019embedding,wallace2014humans,joshi2015harnessing,amir2016modelling,oraby-etal-2016-creating,hazarika2018cascade,isarcasm2020}, to date, sarcasm detection was only explored in a general domain (e.g., the general Twitter or general Reddit) with no focus on a specific context, e.g., a disaster context. 

Natural disasters such as hurricanes and earthquakes cause substantial material destruction and emotional damage to millions of people every year \cite{ritchie2014natural}, with many being uprooted, evacuated, or disrupted from their daily activities. During disasters, affected individuals often turn to social
media platforms such as Twitter and Facebook to look for updates, to post requests for help, to share their feelings, and to express contempt towards the unfolding event or public policies and guidelines. This contempt is in some cases expressed as the sophisticated linguistic phenomenon 
that makes use of figurative language: the sarcasm (or irony).
Understanding this form of speech in a disaster-centric context is essential to improving 
the understanding of disaster-related tweets and their intended semantic meaning.
However, detecting sarcasm solely from disaster-related short tweets when no visual or acoustic information is available is very challenging because of the difficulty of the task itself even for humans (in the absence of the writer's intent); the lack of sufficient textual context to leverage on; and the lack of annotated datasets for this task. 

To this end, we explore the difficulty of sarcasm detection in disaster-related tweets and
present \textsc{HurricaneSARC}, a dataset that contains $15,000$ tweets annotated with \emph{sarcasm} and \emph{non-sarcasm} using crowdsourcing. Unlike existing sarcasm datasets that cover a general domain, to our
knowledge, \textsc{HurricaneSARC} is the first dataset labeled for sarcasm that covers a specific domain, i.e., tweets sampled from a disaster-centric domain---Hurricanes Harvey, Irma, and Maria, with the goal to measure progress on sarcasm detection in a focused context.  
Table \ref{tab:dataset_samples} shows examples of sarcastic and non-sarcastic tweets from \textsc{HurricaneSARC}. As we can see, in the tweet \emph{Are you having a good laugh in your dry, safe, comfortable home while \#Harvey destroys lives.}, the writer is expressing sarcasm implicitly by using conceptually opposite phrases, such as \emph{good laugh}, \emph{safe}, and \emph{Harvey destroys lives}. On the other hand, in the tweet \emph{Sending good vibes to people in Puerto Rico!!  stay safe.}, the writer is expressing sympathy towards the individuals affected by the disaster, hence it contains no sarcastic information.

Using \textsc{HurricaneSARC}, we contrast BERTweet \cite{bertweet}, a pre-trained language model that learns effective language representations from unlabeled Twitter data, with Convolutional Neural Networks  \cite{kim-2014-convolutional} and BiLSTM \cite{hochreiter1997long} to establish a baseline on our dataset. 

Moreover, we improve upon this baseline result, by leveraging intermediate task pre-training \cite{pruksachatkun-etal-2020-intermediate} to investigate information transfers between sarcasm detection and related tasks such as emotion and sentiment analysis.

\begin{table}[t]
\small
\centering
\begin{tabular}{p{2.5cm}p{4.5cm}}
\toprule
\textsc{sarcastic} & Are you having a good laugh in your dry, safe, comfortable home while \#Harvey destroys lives. \\
\midrule
\textsc{sarcastic} & We know natural disasters don't discriminate, but relief and who is most severely impacted tend to do so. \#Harvey. \\
\midrule
\textsc{sarcastic} & \#Trump is excited by 125 mph winds of \#Hurricane \#Harvey in the way that a child is excited by his toys! Disgusting.\\
\midrule
\textsc{not sarcastic} & Sending good vibes to people in Puerto Rico!!  stay safe. \\
\midrule
\textsc{not sarcastic} & \#Harvey made landfall in Tex. at 11p as first major (Cat 3+) hurricane since Wilma in 2005. \\
\midrule
\textsc{not sarcastic} & Maria, although downgraded to a category 4, is still a monster of a storm tearing through Puerto Rico right now. \\
\bottomrule
\end{tabular}
\caption{Examples from our dataset.}
\label{tab:dataset_samples}
\end{table}

As disasters strike, large amounts of user-generated content are produced on social sites. However, due to the nature of disasters unfolding rapidly and the high costs needed for annotation, only a small quantity can be annotated and used for supervised classification. To address these issues, in this paper we investigate and contrast two potential solutions. First, we carry out experiments to study models' understanding of sarcasm in disaster-related tweets when trained on existing datasets for sarcasm from the general domain (direct transfer). Second, we propose semi-supervised learning (SSL) to exploit the readily available unlabeled data together with small labeled data.
Specifically, we explore \emph{Uncertainty-aware Self-Training (UST)} \cite{mukherjee2020uncertainty} and \emph{Noisy Student Training} \cite{xie2020self} using noising techniques such as \emph{back translation} \cite{sennrich2015improving}, and show that unlabeled data can improve the performance of our models significantly. Concretely, our BERTweet-based \emph{UST} model that uses only $200$ labeled examples performs within $1\%$ F1 of the vanilla BERTweet model, which is trained on four times more data. In contrast, transfering information from domains such as Twitter or Reddit is not as effective as our SSL methods, decreasing the overall F1 by $15\%$.

We summarize our contributifons as follows: 1) We introduce \textsc{HurricaneSARC}, a new dataset for sarcasm detection in the disaster domain and show that models trained on a general domain struggle on sarcasm detection when evaluated on a disaster domain; 2) We develop strong baselines on our dataset, and explore intermediate task transfer learning as a means to improve the performance on our dataset; 3) We propose semi-supervised learning as a means to obtain better language representations for time-critical events such as disasters.  

\section{Related Work}

Due to its highly figurative nature, identifying sarcasm from textual data is defined as one of the most challenging tasks in NLP  \cite{wallace2014humans}, which has attracted significant attention in recent years.

The approaches proposed range from rule-based methods \cite{veale2010detecting,maynard-greenwood-2014-cares,bharti2015parsing,singh2018footwear,riloff2013sarcasm} and statistical approaches \cite{joshi2015harnessing,tepperman2006yeah,kreuz2007lexical,reyes2012making,liebrecht2013perfect, wallace2015sparse,lukin-walker-2013-really,oraby-etal-2016-creating}, to deep learning techniques \cite{joshi2016word,amir2016modelling,ghosh2016fracking,hazarika2018cascade,DBLP:journals/corr/abs-1901-08014,kumari2017parallelization}. 

\citet{maynard-greenwood-2014-cares} argued that the sentiment hashtags can be key indicators for the expression of sarcasm, and developed rule-based classifiers, aimed at identifying negative phrases in positive sentences, as well as mining hyperboles, which are usually great indicators of sarcasm. \citet{reyes2012making} used Na\"ive Bayes and Decision Trees to identify sarcasm in tweets based on features such as the presence of irony or humor.
Other studies focused on recognizing  
interjections, punctuation symbols, intensifiers, hyperboles \cite{kreuz2007lexical}, emoticons \cite{carvalho2009clues}, exclamations \cite{tsur2010icwsm,10.1007/978-981-10-8639-7_25}, and hashtags \cite{davidov2010semi} in sarcastic comments.
On the other hand, \citet{ghosh2016fracking} proposed a concatenation of Convolutional Neural Network, Long-Short Term Memory Network, and Deep Neural Network (CNN-LSTM-DNN) that outperformed many state-of-the-art statistical methods based on text features. \citet{poria2016deeper} developed a framework based on a pre-trained CNN to retrieve sentiment, emotion, and personality features for sarcasm recognition. \citet{zhang2016tweet} used a bi-directional gated recurrent neural network with a pooling mechanism to automatically detect features from tweets, as well as context information from the history of the author. 

\citet{DBLP:journals/corr/abs-1901-08014} developed a multi-task learning framework and applied it on a dataset of $1,000$ sentences annotated with both sarcasm and sentiment tags, introduced by \citet{mishra2016predicting}. 
Their framework is based on a GRU \cite{chung2014empirical} model, and fuses the sarcasm and sentiment-specific vectors through a tensor network.


In addition to developing approaches for identifying sarcasm, researchers also focused on creating datasets from different online platforms.
SARC \cite{DBLP:journals/corr/KhodakSV17} is a general Reddit Corpus encompassing 1.3 million sarcastic statements. The dataset is self-annotated (i.e., the sarcasm label is assigned by the author of the statement). iSarcasm \cite{isarcasm2020} is a dataset of intended sarcasm from the general Twitter domain, which consists of 4,484 tweets manually labeled with the \emph{sarcastic} and \emph{non-sarcastic} categories. Several other datasets exist for sarcasm detection from platforms such as general Twitter and Reddit that are annotated using crowdsourcing \cite{riloff2013sarcasm,maynard-greenwood-2014-cares,ptavcek2014sarcasm,oraby-etal-2016-creating}, or leveraging platform specific cues such as hashtags in Twitter \cite{davidov2010semi,gonzalez2011identifying,reyes2013multidimensional}. For example, \citet{riloff2013sarcasm} annotated $1,600$ tweets from the general Twitter domain with the \emph{sarcastic} and \emph{non-sarcastic} categories using crowdsourcing. On the other hand, \citet{davidov2010semi} used hashtags such as \#Sarcastic or \#Sarcasm to create a corpus composed of $900$ examples annotated with three categories: \emph{positive}, \emph{negative}, and \emph{sarcastic}.

In contrast, \textsc{HurricaneSARC} contains tweets streamed during natural disaster events. To our knowledge, no previous dataset for sarcasm detection covers a specialized domain, e.g., a disaster domain. Moreover, composed of $15,000$ tweets manually annotated for \emph{sarcasm}, our dataset enables complex exploration of pre-trained language models such as BERT \cite{devlin-etal-2019-bert}, as well as the study of domain gaps between our disaster setting and other general domains. We hope that \textsc{HurricaneSARC} will spur further research and offer a better understanding of time-critical and crucial events such as disasters.

\vspace{-1mm}
\section{Dataset}
\label{section:dataset}
\noindent
In this section, first we detail the construction of HurricaneSARC, followed by our anlysis into the lexical particularities of our dataset.

\subsection{Dataset Construction} 

\textsc{HurricaneSARC} is a dataset of English tweets from the disaster domain, annotated for sarcasm. We randomly sampled $15,000$ tweets from the large-scale repository of tweets introduced by \citet{raychowdhury2019keyphrase}, that were streamed during the Hurricanes Irma, Harvey, and Maria. Next, we labeled these tweets using the Amazon Mechanical Turk crowdsourcing platform. The Turk workers were provided definitions of sarcasm from dictionary.com and Wikipedia and were asked to annotate tweets for the expression of sarcasm, i.e., to assign the \emph{true} label for tweets perceived as containing sarcasm (or irony), and \emph{false} otherwise. The annotators were paid fairly. Each tweet was annotated by five different workers, and the final label for a tweet was computed using majority vote. To ensure qualitative results, we employed strict annotator requirements for the task, such as high acceptance rate (>$95$\%), and a large amount of completed tasks ($500+$).

The annotation process produced $820$ sarcastic tweets, which amount for $5.5\%$ of the total number of sampled tweets. 
These tweets represent positive samples (i.e., tweets labeled as expressing sarcasm). To create the negative samples, we sampled an equal amount from the remaining pool (i.e., the tweets labeled as not expressing sarcasm). We also experimented with all negative samples as opposed to downsampling an equal amount. However, this significantly hurt model performance. Thus, consistent with \citet{DBLP:journals/corr/KhodakSV17}, we construct a balanced version of the dataset on which we report the modeling results in the following sections. However, we make both dataset versions publicly available to enable further progress on sarcasm detection and 
understanding of sarcasm in a focused context. 
In our final balanced version of the dataset, we use \textasciitilde$25\%$ of the data for testing, \textasciitilde$18\%$ for validation, and the rest for training. Table \ref{number_examples} shows the number of examples in each split.

\begin{table}[h]
\small
\centering
\resizebox{\columnwidth}!{
\begin{tabular}{lccc}
\toprule
\textbf{Set} & \textbf{Sarcastic} & \textbf{Non-Sarcastic} & \textbf{Total} \\
\midrule
Train & $467$ & $467$ & $934$ \\ 
Validation & $150$ & $150$ & $300$ \\ 
Test & $200$  & $200$ & $400$ \\
\hline
\end{tabular}
}
\caption{\textsc{HurricaneSARC} benchmark dataset size.}
\label{number_examples}
\end{table}

\begin{table*}[h]
\small
\setlength{\tabcolsep}{8pt}
\centering
\begin{tabular}{r|cc|ccccccc}
\toprule
& \textsc{pos} & \textsc{neg} & \textsc{joy} & \textsc{fer} & \textsc{sdn} & \textsc{dsg} & \textsc{ang}  & \textsc{ant} & \textsc{srp} \\
\midrule
\textsc{sarcastic-hs} & $0.48$ & $0.70$ & $0.18$ & $0.57$ & $0.28$ & $0.22$ & $0.32$ & $0.23$ & $0.25$ \\ 
\textsc{non-sarcastic-hs} & $0.41$ & $0.58$ & $0.34$ & $0.52$ & $0.20$ & $0.13$ & $0.26$ & $0.22$ & $0.20$ \\ 
\midrule
\textsc{sarcastic-is} & $0.53$ & $0.35$ & $0.34$ & $0.21$ & $0.18$ & $0.12$ & $0.19$ & $0.35$ & $0.19$\\ 
\textsc{non-sarcastic-is} & $0.49$ & $0.33$ & $0.32$ & $0.19$ & $0.20$ & $0.15$ & $0.18$ & $0.33$ & $0.18$\\ 
\hline
\end{tabular}
\caption{Co-occurence of sarcastic/non-sarcastic tweets from \textsc{HurricaneSARC} (HS) and  \textbf{iSarcasm} (IS) with words expressing a positive (POS) or negative (NEG) sentiment, as well as words expressing emotions such as joy (JOY), fear (FER), sadness (SDN), disgust (DSG) and anger (ANG), anticipation (ANT) and surprise (SRP). The numbers are normalized by the number of tweets in the sarcastic or non-sarcastic class.}
\label{emo_co_occurence}
\end{table*}

\begin{table}[t]
\small
\centering
\begin{tabular}{p{2.3cm}p{4.3cm}}
\toprule
HurricaneSARC & My neighbor is giving hurricane tips on Facebook meanwhile he already evacuated and left his yard full of projectiles. \\
\midrule
HurricaneSARC & Holy Christ, you insufferable git. How about you help Puerto Rico? They were directly hit by a hurricane. Remember?. \#Harvey. \\
\midrule
iSarcasm & talents: making a joke of myself by crying in front of class.\\
\midrule
iSarcasm & "healthcare is not a right"... Ok \\
\bottomrule
\end{tabular}
\caption{Examples from the \textsc{HurricaneSARC} and \textbf{iSarcasm} dataset.}
\label{tab:isarc_vs_hsarc}
\end{table}

\begin{table}[h]
\small
\centering
\begin{tabular}{lcc}
\toprule
\textbf{Dataset} & \textbf{Sarcastic} & \textbf{Non-Sarcastic} \\
\midrule
HurricaneSARC & $23.1$ & $19.3$ \\ 
iSarcasm & $17.2$ & $18.5$ \\ 
\hline
\end{tabular}
\caption{Average number of words per tweet for sarcastic/non-sarcastic comments.}
\label{number_words}
\vspace{-3mm}
\end{table}

\subsection{Lexical Analysis}

 We perform a word level lexical analysis to gain insights into the vocabulary used in \textsc{HurricaneSARC} (compared with a general domain) and explore potential challenges and particularities of our dataset. First, we investigate the occurence of emotion and sentiment intensive words in sarcastic and non-sarcastic tweets.

To this end, we use EmoLex \cite{mohammad2013crowdsourcing}, a lexicon containing words and the associated emotion and sentiment evoked by these words. We study this co-occurence on \textsc{HurricaneSarc} by contrast with the iSarcasm dataset introduced by \citet{isarcasm2020} from a general domain and report the co-occurence ratios in Table \ref{emo_co_occurence}. Note that both datasets are compiled from Twitter. First, we observe that only $18\%$ of the sarcastic tweets in \textsc{HurricaneSarc} contain words expressing joy, whereas the ratio of non-sarcastic tweets containing joy words is significantly larger, up to $34\%$. In contrast, we observe a different pattern in the iSarcasm dataset, where both the sarcastic and non-sarcastic tweets contain a similar ratio of words expressing joy. On the same note, the \textsc{HurricaneSARC} dataset has a considerable amount of negatively polarized words. In fact, as many as $70\%$ of the tweets in \textsc{HurricaneSARC} contain at least a negative word, whereas in iSarcasm, this ratio is halved at $35\%$.

These findings suggest that the disaster-related tweets tend to be more emotion and sentiment-intensive than tweets from the general Twitter domain, since there are higher stakes involved in disaster scenarios. To better understand the particularities of our \textsc{HurricaneSARC} dataset, we compare a few representative examples from our dataset with some representative examples from the iSarcasm dataset in Table \ref{tab:isarc_vs_hsarc}. We observe that \textsc{HurricaneSARC} is focused on the event and sarcasm is conveyed towards the ongoing disaster, while iSarcasm is much broader, containing tweets where the sarcasm is expressed towards various targets such as self-deprecation or healthcare.

We also note differences in the average number of words per tweet in our disaster scenario compared to the general Twitter domain. We show our findings in Table \ref{number_words}, where we compare our disaster setting with the general Twitter domain from the iSarcasm dataset \cite{isarcasm2020}. Interestingly, our  tweets are longer in length than previous datasets, with the sarcastic tweets containing six additional words on average compared to iSarcasm. 
\section{Baseline Modeling} 

We model the sarcasm detection on \textsc{HurricaneSARC} using various methods:

 \paragraph{Traditional Neural Methods} We experiment with \textbf{(1) CNN} for text classification introduced by \citet{kim2014convolutional} and \textbf{(2) Bi-LSTM} \cite{hochreiter1997long} using pre-trained GloVe \cite{pennington2014glove} word embeddings as inputs.
 
\paragraph{Pre-trained Language Models} We use the BERTweet \cite{bertweet} large-scale pre-trained language model for English Tweets, on top of which we add a linear layer for classification. We also experiment with the vanilla BERT model \cite{devlin-etal-2019-bert}.

\subsection{Experimental Setting}

All models are implemented using the Huggingface Transformers \cite{wolf-etal-2020-transformers} library. For training, we use a single Nvidia V$100$ GPU. We detail the best hyperparameters used and the computational resources needed to train our models in Appendix A. To ensure statistically significant results, we run the experiments $5$ times, with different model weight initialization, and report the average of the results obtained from the five independent runs.

\subsection{Results}

We show the results using the above baselines in Table \ref{tab:model_performance}. First,  we observe that CNN performs slightly better than BiLSTM. Second, we notice that the BERTweet model pre-trained on a Twitter corpus outperforms the vanilla BERT model trained on Wikipedia and Bookcorpus by $1.7\%$ in F1, and the other baselines by as much as $9\%$ in F1. These results show that pre-trained language models are successful in learning useful language representations that are beneficial for the detection of sarcasm, and platform-specific pre-training (BERTweet pre-training) further improves the performance (compared with vanilla BERT).

Overall, despite that BERTweet performs better than the other baselines, its performance (in-domain) is still low, with the best method achieving only $0.702$ F1 (Table \ref{tab:model_performance}). 
Next, we investigate various ways to introduce inductive biases into our models by performing a thorough investigation into intermediate task transfer learning.

\begin{table}[t]
\small
\setlength{\tabcolsep}{10pt}
\centering
\begin{tabular}{r|ccc}
\toprule
\textsc{method} & \textsc{p} & \textsc{r} & \textsc{f1} \\ 
\midrule
\textsc{Bi-LSTM} & $0.595$ & $0.610$ & $0.600$  \\
\textsc{CNN} & $0.625$ & $0.613$ & $0.613$  \\
\textsc{BERT} & $0.665$ & $0.702$ & $0.685$ \\
\textsc{BERTweet} & $\bf{0.692}$ & $\bf{0.713}$ & $\bf{0.702}$ \\
\bottomrule
\end{tabular}

\caption{Precision (P), recall (R), F1-score (F1) on \textsc{HurricaneSARC} using neural models. We assert significance$^{\dagger}$ if $p < 0.05$ under a paired-t test.}
\label{tab:model_performance}
\vspace{-3mm}
\end{table}

\begin{table*}[h]
\vspace{-4mm}
\small
\setlength{\tabcolsep}{1pt}
\centering

\begin{tabular}{r|ccccc|ccc}
\toprule
\textsc{domain} & \textsc{carer} & \textsc{emonet} & \textsc{isarcasm} & \textsc{goemotions} & \textsc{sarc} & \textsc{emonet} & \textsc{sarc} & \textsc{hurricane-ext}\\
\midrule
& \multicolumn{5}{c}{\textsc{supervised}} & \multicolumn{3}{c}{\textsc{unsupervised}} \\
\midrule
No. examples & $50,000$ & $50,000$ & $4,484$ & $58,000$ & $50,000$ & $50,000$ & $1.3M$ & $15M$ \\ 

\bottomrule
\end{tabular}
\caption{Number of examples in each intermediate task.}
\label{transfer_learning}
\vspace{-4mm}
\end{table*}

\section{Intermediate-Task Transfer Learning}
\label{section:intermediate}

To improve the performance of our baseline BERTweet model, we carry out a comprehensive set of experiments using intermediate task transfer learning \cite{han-eisenstein-2019-unsupervised,pruksachatkun-etal-2020-intermediate,Gururangan2020DontSP} (i.e., adjusting the contextualized embeddings to a target domain). Our framework is composed of two steps: First, we pre-train the BERTweet model on an intermediate supervised or unsupervised task. In the supervised setting, we add a task-specific classification layer, then train the model using a supervised loss on the intermediate task. On the other hand, in the unsupervised setting, we pre-train on unlabeled corpus using the dynamic masked language modeling objective (MLM). Second, after we train our models on an intermediate task, we remove any intermediate-task classification layers and initialize a new linear layer on top of BERTweet for training the model on \textsc{HurricaneSARC}.

\subsection{Supervised Pre-training}
We explore five intermediate tasks (described below) for the supervised scenario: 1) {\em EmoNet} is a dataset \cite{abdul-mageed-ungar-2017-emonet} for fine-grained emotion detection in the general Twitter domain. It contains $1.6$ million tweets labeled with the Plutchik-$8$ basic emotion categories \cite{plutchik1980general}. Here, we use a smaller version of the dataset containing $50,000$ examples provided by the authors; 2) {\em EmoNetSent} uses the same data as \emph{EmoNet}. However, we aggregate the emotion labels based on the positive, negative, and neutral polarities; 3) {\em GoEmotions} \cite{goemotions2020} is a dataset of $58,000$ Reddit comments annotated with $27$ types of emotions; 4) {\em CARER} \cite{saravia-etal-2018-carer} contains 664,462 English tweets from the general Twitter domain annotated with 8 emotion categories. To ensure a fair comparison with the previous datasets, we downsample the data to $50,000$ examples and use only these in our intermediate pre-training step; 5) {\em SARC} \cite{DBLP:journals/corr/KhodakSV17} and 6) {\em iSarcasm} \cite{isarcasm2020} are the datasets presented earlier in the paper. 

\vspace{-3mm}
\subsection{Unsupervised Pre-training}
For unsupervised intermediate tasks, we pre-train on dynamic masked language modeling (MLM) on three corpora: 1) {\em Hurricane} \cite{raychowdhury2019keyphrase} is a large-scale repository of $15$M tweets streamed during the Hurricanes Irma, Harvey, and Maria; 2) {\em EmoNet} \cite{abdul-mageed-ungar-2017-emonet} is the emotion corpus from Twitter presented earlier. However, we remove the labels and train only on the tweets; 3) {\em SARC} is the sarcasm Reddit corpus presented earlier. Similar to \emph{EmoNet}, we remove the labels and train only on the comments.
The above datasets are summarized in Table \ref{transfer_learning}.

\begin{table}[t]
\small
\setlength{\tabcolsep}{5pt}
\centering
\begin{tabular}{r|ccc}
\toprule
\textsc{method} & \textsc{p} & \textsc{r} & \textsc{F1} \\ 
\midrule
\textsc{BERTweet} & $0.692$ & $0.713$ & $0.702$ \\
\midrule
\textsc{carer} & $0.715$ & $0.712$ & $0.713$  \\
\textsc{emonet} & $0.685$ & $0.793$ & $\bf{0.732}$$^{\dagger}$ \\
\textsc{emonet sent} & $0.684$ & $0.653$ & $0.662$ \\
\textsc{isarcasm} & $0.645$ & $\bf{0.803}$ & $0.712$ \\
\textsc{goemotions} & $0.632$ & $0.703$ & $0.665$ \\
\textsc{sarc} & $0.652$ & $0.713$ & $0.687$ \\
\midrule
\textsc{emonet} & $0.652$ & $0.683$ & $0.675$ \\
\textsc{sarc} & $0.684$ & $0.707$ & $0.698$ \\
\textsc{hurricane-ext} & $0.714$ & $0.731$ & $0.723$$^{\dagger}$ \\
\bottomrule
\end{tabular}
\caption{Precision, recall, and F1 using the following methods: \textbf{1)} Supervised Intermediate Pre-training using BERTweet (upper block). \textbf{2)} Unsupervised Pre-training using BERTweet (lower block). We assert significance$^{\dagger}$ if $p < 0.05$ under a paired-t test with vanilla BERTweet.}
\label{tab:model_performance_transfer}
\end{table}

\subsection{Results} Our intermediate task pre-training experiment results, shown in Table \ref{tab:model_performance_transfer}, reveal interesting details: First, supervised pre-training on the \emph{EmoNet} emotion detection dataset is able to improve the performance over the simple BERTweet model by as much as $3\%$ in F1. Meanwhile, we also notice a $1\%$ improvement when using the \emph{iSarcasm} dataset, as well as a $1\%$ improvement using the \emph{Carer} task. All these tasks contain data from the Twitter domain, which indicates that positive transfers of information tend to occur when transferring information between two related domains. In contrast, we observe negative transfers of information between \emph{SARC} or \emph{GoEmotions} and \textsc{HurricaneSARC}. We attribute the performance decrease to the disparity between the domains and platforms, i.e., both \emph{SARC} and \emph{GoEmotions} are collected from the Reddit platform, whereas our target domain is Twitter. In the unsupervised settings, performing MLM on \textsc{Hurricane} yields the best performance improvement of $2$\% in F1. Interestingly, unlike the supervised scenario, pre-training on \emph{SARC} is able to improve the performance over the vanilla BERTweet model, while \emph{EmoNet} decreases the F1 score. However, both \textsc{Hurricane} and \emph{SARC} underwent considerably more pre-training than \emph{EmoNet}, due to the significantly smaller size of \emph{EmoNet}. 

We summarize our results as follows: 1) Both emotion detection and sarcasm detection make good supervised intermediate tasks for sarcasm detection, as long as the data used originates from the same social platform; 2) Additional unsupervised pretraining using masked language modeling helps the performance even when the intermediate data differs from the target data, provided that the tasks are closely related (e.g., sarcasm domain) and the training data has a considerable size. We also provide results of the experiments using the vanilla BERT model in Appendix B.

\begin{table}[t]
\small
\centering
\begin{tabular}{p{2.5cm}p{4.4cm}}
\toprule
\textsc{Sarcastic} & So you are happy about the death and destruction about to be done in Florida by IRMA ? I hope one of those. \\
\midrule
\textsc{Sarcastic} & People are dying in life support in Puerto Rico and the world is still in a frenzy about football. \\
\midrule
\textsc{Not Sarcastic} & Maria-ravaged Puerto Rico left in misery without power, water and food\\
\midrule
\textsc{Not Sarcastic} & If you're angry about the NFL then you need to get over yourselves because 3.5 MILLION AMERICANS are at risk of DEATH in Puerto Rico \\
\bottomrule
\end{tabular}
\caption{Examples corrected by the \emph{EmoNet} BERTweet model.}
\label{tab:error_analysis}
\vspace{-4mm}
\end{table}

\begin{table}[t]
\small
\centering
\resizebox{\columnwidth}!{
\begin{tabular}{lccc}
\toprule
\backslashbox{Train}{Test} & SARC & iSarcasm & HurricaneSARC \\
\midrule
SARC & $0.761$ & $0.342$ & $0.513$ \\ 
iSarcasm & $0.672$ & $0.391$ & $0.552$\\ 
HurricaneSARC & $0.551$ & $0.286$ & $0.702$\\ 
\hline
\end{tabular}
}
\caption{F-1 scores of BERTweet when trained and tested on different sarcasm datasets.}
\label{domain_adaptation}
\vspace{-4mm}
\end{table}

\subsection{Error Analysis} To investigate the effect of intermediate task fine-tuning, we perform an error analysis of our models. Specifically, we pass the examples from the test set through two models. First, we get the predictions of the vanilla BERTweet model, which underwent no additional pre-training. Second, we compute the predictions of the best performing emotion-aware BERTweet model, which underwent additional supervised pre-training on the \emph{EmoNet} emotion dataset. Finally, we analyze examples where additional pre-training is able to change incorrect predictions into correct predictions, or correct predictions into incorrect ones. Our \emph{EmoNet} BERTweet misclassifies $4$ examples that are correctly classified by the vanilla BERTweet model. On the other hand, there are $28$ examples misclassified by the simple BERTweet model, which are correctly classified by the \emph{EmoNet} BERTweet model. We show a few of these examples in Table \ref{tab:error_analysis}. Interestingly, these examples convey strong emotional messages. For example, in \emph{People are dying in life support in Puerto Rico and the world is still in a frenzy about football.}, sadness and anger are strongly expressed along sarcasm. We argue that the supervised pre-training on \emph{EmoNet} is able to induce important information about the emotions in a tweet, which correlate with the expressed sarcasm.

\begin{table*}[t]
\vspace{-6mm}
\small
\setlength{\tabcolsep}{18pt}
\centering
\begin{tabular}{r|ccc}
\toprule
\textsc{method}  & \textsc{p} & \textsc{r} & \textsc{F1} \\
\midrule
\textsc{bertweet-$100$} & $0.591$ & $0.582$ & $0.589\pm2.7$\\
\textsc{bertweet-$200$} & $0.651$ & $0.633$ & $0.642\pm2.0$\\
\textsc{bertweet-all} & $0.692$ & $0.713$ & $0.702\pm0.94$ \\
\midrule
\textsc{noisy student-$100$} & $0.628$ & $0.612$ & $0.620\pm2.3$ \\
\textsc{noisy student-$200$} & $0.671$ & $0.673$ & $0.672\pm1.7$ \\
\textsc{noisy student-all} & $0.737$ & $0.735$ & $0.736^{\dagger}\pm1.2$ \\
\midrule
\textsc{ust-$100$} & $0.633$ & $0.622$ & $0.631\pm2.2$ \\
\textsc{ust-$200$} & $0.691$ & $0.686$ & $0.689^{\dagger}\pm1.7$ \\
\textsc{ust-all} & $\bf{0.759}$ & $\bf{0.752}$ & $\bf{0.755\pm0.69^{\dagger}}$ \\
\bottomrule
\end{tabular}
\caption{Average precision, recall, and F-1 (averaged across the $5$ runs) using semi-supervised learning and the vanilla BERTweet model. The number following the model name indicates the number of examples that model was trained on (e.g., BT-$30$ is the vanilla BERTweet model trained on $30$ labeled examples). We assert significance$^{\dagger}$ if $p < 0.05$ under a t-test with the counterpart vanilla BERTweet model (e.g., UST-$200$ vs. BT-$200$).}
\label{tab:model_performance_semisupervised}
\vspace{-3mm}
\end{table*}

\section{Dealing with Time-Critical Events}

As disasters start to unfold and unlabeled data starts to accumulate rapidly, annotating large amounts of tweets for sarcarm becomes very challenging. The time-critical nature of disasters and the annotation costs are two of the main challenges to understanding an ongoing disaster. In this section, we investigate various methods to overcome these challenges. We ask the following questions: Does one need annotated data at all to detect perceived sarcasm in an emerging disaster? If annotated data is vital, can we leverage only a very small set of labeled examples to obtain good performance instead?

\subsection{Direct Transfer}

To answer the first question, we stage the following setup: We train our best BERTweet model on out-of-domain sarcasm detection datasets, then \emph{directly} test the model on \textsc{HurricaneSARC}. We experiment with iSarcasm \cite{isarcasm2020} and SARC \cite{DBLP:journals/corr/KhodakSV17} as our out-of-domain datasets, which both cover a general domain (Twitter and Reddit, respectively), whereas \textsc{HurricaneSARC} covers a specialized domain (Twitter). Since SARC is much larger in size compared with the other datasets, 
we downsample SARC to $5,000$ examples (which is roughly the size of iSarcasm) to discount for the impact of dataset size. We show the results of these experiments in Table \ref{domain_adaptation} and make a few observations. First, training on SARC  or iSarcasm  and testing on \textsc{HurricaneSARC} hurts the performance considerably, decreasing the F1 by $18.9\%$ and $15\%$, respectively, compared with training and testing on \textsc{HurricaneSARC}. It is interesting to see that the drop in performance is much higher when training on Reddit compared with Twitter, which suggests that differences exist in the way sarcasm is expressed across platforms. On the other hand, training on iSarcasm and testing on SARC shows a decrease of only $9\%$ in F1, whereas we note a decrease of $21\%$ in F1 when training on \textsc{HurricaneSARC} and testing on SARC. These results reinforce that in-domain labeled data is vital in providing good model performance and understanding of the ongoing disaster. However, given the rapid unfolding of a disaster and considering that data annotation is a time-intensive process, only a small amount of annotations can be obtained as the disaster starts to unfold. 

Therefore, we now turn to our second question, and propose semi-supervised learning approaches as the solution.

\subsection{Semi-supervised learning}
\label{section:ssl}

Our proposed methods can use large amounts of unlabeled data generated during natural disasters to considerably reduce the annotation costs as well as reduce the time needed to acquire the labeled data. We explore two state-of-the-art semi-supervised techniques, and train these methods on small subsets of our training set. Using a quarter of the entire training set, these methods perform similarly with techniques leveraging the whole set of labeled examples. We consider the following methods:

\textbf{1) Noisy student training} \cite{xie2020self}  is a approach leveraging knowledge distillation and self-training, which iteratively jointly trains two models in a teacher-student framework. Noisy student uses a larger model size and noised inputs, exposing the student to more difficult learning environments, which usually leads to an increased performance compared to the teacher. To add noise to our input examples, we use two approaches: a) {\em Synonym replacement:} We replace between one and three words in a tweet with its synonym using the WordNet English lexical database \cite{fellbaum2012wordnet}; b) {\em Back-translation:} We use {back-translation}, and experiment with different levels of noise corresponding to different translation chain lengths (e.g., English-French-Spanish-English). Smaller chain lengths lead to less noise, while increasing the length of the chain produces examples with significantly more noise. \textbf{2) Uncertainty aware Self-Training} \cite{mukherjee2020uncertainty} incorporates uncertainty estimates into the standard \emph{teacher-student} self-training framework by adding a few highly effective changes to the typical teacher-student self-training framework, such as acquisition functions using Monte Carlo dropout and a new learning mechanism leveraging teacher model confidence.

\subsection{Reducing Time \& Annotation Costs} We investigate how the above semi-supervised learning approaches can help  reduce the human effort needed to annotate the data. To this end, we train the  models on subsets of our dataset with only $100$ and with only $200$ training examples per class. These $100$- and $200$-size subsets are created randomly. We run each of these methods with $10$ different randomly chosen subsets, and report the mean and standard deviation of the obtained F1 for each model.

\subsection{Results} We show the results of the mentioned approaches in Table \ref{tab:model_performance_semisupervised}. We make the following observations. Our experiments involving $200$ (UST-$200$) training examples per class show the feasibility of using semi-supervised approaches for sarcasm detection. Using as little as $25\%$ of the available data, semi-supervised approaches perform similar to models trained on the whole dataset. Therefore, the semi-supervised approaches can deliver the same performance but use four times less annotated data. Since disasters are time-critical events, these results show that SSL methods can significantly contribute to a faster and more reliable understanding of an unfolding disaster. Moreover, the F1 variance of different runs of all the UST (UST-100, UST-200, and UST-ALL) methods is significantly improved compared to the vanilla BERTweet, which shows that semi-supervised models are considerably more stable.

\section{Conclusion}
In this paper, we introduced \textsc{HurricaneSARC}, a dataset for perceived sarcasm detection,  composed of $15,000$ English tweets from multiple hurricane events, and annotated with the \emph{sarcastic} and \emph{non-sarcastic} labels. We detailed the potential particularities and challenges of our dataset, and developed neural baselines for \textsc{HurricaneSARC}. Next, we improved the performance of our pre-trained language models using supervised or unsupervised intermediate task transfer learning, as well as semi-supervised learning techniques. The best BERTweet model (UST-ALL) is able to improve the performance of our baselines by $5.3\%$ F-1 score, by leveraging the large amounts of unlabeled data from the disaster domain. 
We hope that our dataset for sarcasm detection in a specific domain will spur research in this area and will lead to novel approaches for in-domain and out-of-domain explorations that will improve natural language understanding of disaster-related tweets as well as the overall understanding of sarcasm. Furthermore, the construction of other specialized datasets, e.g., from a financial domain, is another interesting direction.  

\section*{Limitations} Our work introduces a new dataset for sarcasm detection from hurricane disasters. While the domain of disasters can enable meaningful applications in the real-world, the label distribution in HurricaneSARC is extremely skewed: negative examples are much more prevalent. To this end, detecting sarcasm is particularly challenging in our context. Moreover, since our data is collected from Twitter, it contains the inherent biases that are rooted into this type of data.

\bibliography{anthology,custom}
\bibliographystyle{acl_natbib}

\newpage

\appendix

\section{Hyperparameters}

We show the hyperparameters used for our best models in Table \ref{resources}. For the \emph{UST} learning rates in scarce label scenarios, we show the range of the learning rates used since different training sets produce different best learning rates. Moreover, for \emph{UST}, we show both the labeled batch size (L), as well as the unlabeled batch size (U). To perform the tuning, we use the Ray tune library \cite{liaw2018tune}. The semi-supervised experiments took $1$ week on a single V$100$ GPU.

\begin{table}[h]
\centering
\small
\begin{tabular}{r|ccc}
\toprule
\textsc{model} & \textsc{lr} & \textsc{batch size} & \textsc{epochs}\\
\midrule
\textsc{bi-lstm} & $5\mathrm{e}{-04}$ & $32$ & $4$ \\ 
\textsc{cnn} & $4\mathrm{e}{-04}$ & $32$ & $5$\\ 
\textsc{bert} & $6\mathrm{e}{-05}$ & $16$ & $3$\\ 
\textsc{bertweet} & $5\mathrm{e}{-05}$ & $16$ & $3$\\ 
\textsc{ust-$100$} & $1\mathrm{e}{-05}$$\rightarrow9\mathrm{e}{-05}$ & $4L$$\rightarrow$$16U$ & $25$ \\ 
\textsc{ust-$200$} & $4\mathrm{e}{-05}$$\rightarrow6\mathrm{e}{-05}$ & $8L$$\rightarrow$$32U$ & $23$ \\ 
\textsc{ust-all} & $6\mathrm{e}{-05}$ & $16L$$\rightarrow$$32U$ & $24$ \\ 
\bottomrule
\end{tabular}
\caption{Best hyperparameters.}
\label{resources}
\end{table}

\section{BERT Results}
We show results of the intermediate-task transfer learning and semi-supervised learning experiments using the plain BERT \cite{devlin-etal-2019-bert} model in Table \ref{tab:model_performance_transfer} and Table \ref{tab:model_performance_semisupervised}.

\begin{table}[t]
\vspace{-5mm}
\small
\setlength{\tabcolsep}{10pt}
\centering
\begin{tabular}{r|ccc}
\toprule
\textsc{method} & \textsc{p} & \textsc{r} & \textsc{F1} \\ 
\midrule
\textsc{carer} & $0.683$ & $0.705$ & $0.693$  \\
\textsc{emonet} & $0.679$ & $0.783$ & $0.781$ \\
\textsc{emonet sent} & $0.651$ & $0.653$ & $0.651$ \\
\textsc{isarcasm} & $0.623$ & $0.788$ & $0.701$ \\
\textsc{goemotions} & $0.612$ & $0.622$ & $0.617$ \\
\textsc{sarc} & $0.671$ & $0.713$ & $0.683$ \\
\midrule
\textsc{emonet} & $0.651$ & $0.681$ & $0.673$ \\
\textsc{sarc} & $0.680$ & $0.701$ & $0.694$ \\
\textsc{hurricane-ext} & $0.691$ & $0.718$ & $0.699$ \\
\bottomrule
\end{tabular}
\caption{Precision, recall, and F1 using the following methods: \textbf{1)} Supervised Intermediate Pre-training using BERT (upper block). \textbf{2)} Unsupervised Pre-training using BERT (lower block).}
\label{tab:model_performance_transfer}
\end{table}

\begin{table*}[t]
\vspace{-5mm}
\setlength{\tabcolsep}{18pt}
\centering
\small
\begin{tabular}{r|cccc}
\toprule
\textsc{method}  & \textsc{p} & \textsc{r} & \textsc{F1} & \textsc{stdev}\\
\midrule
\textsc{bert-$100$} & $0.583$ & $0.575$ & $0.584$ & $\pm2.9$\\
\textsc{bert-$200$} & $0.632$ & $0.623$ & $0.625$ & $\pm2.1$\\
\textsc{bert-all} & $0.663$ & $0.707$ & $0.688$ & $\pm0.98$ \\
\midrule
\textsc{noisy student-$100$} & $0.593$ & $0.595$ & $0.598$ & $\pm3.1$ \\
\textsc{noisy student-$200$} & $0.662$ & $0.665$ & $0.663$ & $\pm2.0$ \\
\textsc{noisy student-all} & $0.732$ & $0.734$ & $0.733$$^{\dagger}$ & $\pm1.5$ \\
\midrule
\textsc{ust-$100$} & $0.623$ & $0.602$ & $0.616$ & $\pm2.7$ \\
\textsc{ust-$200$} & $0.683$ & $0.666$ & $0.674$$^{\dagger}$ & $\pm1.9$ \\
\textsc{ust-all} & \textbf{$0.752$} & \textbf{$0.745$} & \textbf{$0.743$}$^{\dagger}$ & \textbf{$\pm0.78$} \\
\bottomrule
\end{tabular}
\caption{Average precision, recall, F-1 and F-1 standard deviation (of the $5$ runs) using semi-supervised learning and the vanilla BERT model. The number following the model name indicates the number of examples that model was trained on (e.g., BERT-$30$ is the vanilla BERT model trained on $30$ labeled examples). We assert significance$^{\dagger}$ if $p < 0.05$ under a paired-t test with the counterpart vanilla BERT model (e.g., UST-$200$ vs. BERT-$200$).}
\label{tab:model_performance_semisupervised}
\end{table*}

\end{document}